\pdfoutput=1

\documentclass[11pt]{article}

\usepackage{ACL}

\usepackage{graphicx}

\usepackage{times}
\usepackage{latexsym}

\usepackage{tcolorbox}

\NewTColorBox{FakepediaExample}{ O{!htbp} m }{%
    floatplacement={#1},
    float,
    width=\columnwidth,
    colback=gray!5!white,
    colframe=gray!75!black,
    title={#2},
}

\DeclareSymbolFont{extraup}{U}{zavm}{m}{n}
\DeclareMathSymbol{\epfl}{\mathalpha}{extraup}{83}
\DeclareMathSymbol{\microsoft}{\mathalpha}{extraup}{81}
\DeclareMathSymbol{\cnrs}{\mathalpha}{extraup}{82}

\usepackage{graphicx}
\usepackage{subcaption}
\usepackage{float}
\usepackage{cancel}

\usepackage[T1]{fontenc}

\usepackage[utf8]{inputenc}

\usepackage{microtype}

\usepackage{inconsolata}

\usepackage[utf8]{inputenc}
\usepackage[T1]{fontenc}
\usepackage{hyphenat}
\usepackage{xspace}
\usepackage{amsmath}
\usepackage{amsfonts}
\usepackage{hyperref}
\usepackage{url}
\usepackage{booktabs}
\usepackage{multirow}
\usepackage{makecell}
\usepackage{caption}
\usepackage{minibox}
\usepackage{bbm}
\usepackage{graphicx}
\usepackage{balance}
\usepackage{mathtools}
\usepackage{color}
\usepackage{marvosym}
\usepackage{ifthen}
\usepackage{textcomp}
\usepackage{enumitem}
\usepackage{verbatim}
\usepackage{algorithm}
\usepackage{numprint}
\usepackage{balance}

\usepackage{amsthm}
\theoremstyle{plain}

\newcommand{\chatoDisplayMode}[1]{#1}

\definecolor{MyRed}{rgb}{0.6,0.0,0.0} 
\definecolor{MyBlack}{rgb}{0.1,0.1,0.1} 
\newcommand{\inred}[1]{{\color{MyRed}\sf\textbf{\textsc{#1}}}}
\newcommand{\frameit}[2]{
  \begin{center}
  {\color{MyRed}
  \framebox[.9\columnwidth][l]{
    \begin{minipage}{.85\columnwidth}
    \inred{#1}: {\sf\color{MyBlack}#2}
    \end{minipage}
  }\\
  }
  \end{center}
}

\newcommand{\note}[2][]{\chatoDisplayMode{\def\@tmpsig{#1}\frameit{{\Pointinghand} Note}{#2\ifx \@tmpsig \@empty \else \mbox{ --\em #1}\fi}}}
\newcommand{\todo}[2][]{\chatoDisplayMode{\def\@tmpsig{#1}\frameit{{\Writinghand} To-do}{#2\ifx \@tmpsig \@empty \else \mbox{ --\em #1}\fi}}}

\newcommand{\abbrevStyle}[1]{#1}

\newcommand{\etc}{\abbrevStyle{etc.}\xspace}

\newcommand{\Secref}[1]{Sec.~\ref{#1}}

\newcommand{\Tabref}[1]{Table~\ref{#1}}
\newcommand{\Figref}[1]{Fig.~\ref{#1}}

\newcommand{\xhdr}[1]{\vspace{1.7mm}\noindent{{\bf #1.}}}

\newcommand{\textcite}[1]{\citeauthor{#1} \shortcite{#1}}

\newcommand{\hide}[1]{}

\makeatletter
\newcommand{\iffont}[2]{\ifthenelse{\equal{\f@family}{#1}}{#2}{}}
\makeatother

  \usepackage{mathptmx}

  \DeclareSymbolFont{greek}{OML}{cmm}{m}{n}
  \DeclareMathSymbol{\alpha}{\mathalpha}{greek}{"0B}
  \DeclareMathSymbol{\beta}{\mathalpha}{greek}{"0C}
  \DeclareMathSymbol{\gamma}{\mathalpha}{greek}{"0D}
  \DeclareMathSymbol{\delta}{\mathalpha}{greek}{"0E}
  \DeclareMathSymbol{\epsilon}{\mathalpha}{greek}{"0F}
  \DeclareMathSymbol{\zeta}{\mathalpha}{greek}{"10}
  \DeclareMathSymbol{\eta}{\mathalpha}{greek}{"11}
  \DeclareMathSymbol{\theta}{\mathalpha}{greek}{"12}
  \DeclareMathSymbol{\iota}{\mathalpha}{greek}{"13}
  \DeclareMathSymbol{\kappa}{\mathalpha}{greek}{"14}
  \DeclareMathSymbol{\lambda}{\mathalpha}{greek}{"15}
  \DeclareMathSymbol{\mu}{\mathalpha}{greek}{"16}
  \DeclareMathSymbol{\nu}{\mathalpha}{greek}{"17}
  \DeclareMathSymbol{\xi}{\mathalpha}{greek}{"18}
  \DeclareMathSymbol{\pi}{\mathalpha}{greek}{"19}
  \DeclareMathSymbol{\rho}{\mathalpha}{greek}{"1A}
  \DeclareMathSymbol{\sigma}{\mathalpha}{greek}{"1B}
  \DeclareMathSymbol{\tau}{\mathalpha}{greek}{"1C}
  \DeclareMathSymbol{\upsilon}{\mathalpha}{greek}{"1D}
  \DeclareMathSymbol{\phi}{\mathalpha}{greek}{"1E}
  \DeclareMathSymbol{\chi}{\mathalpha}{greek}{"1F}
  \DeclareMathSymbol{\psi}{\mathalpha}{greek}{"20}
  \DeclareMathSymbol{\omega}{\mathalpha}{greek}{"21}
  \DeclareMathSymbol{\varepsilon}{\mathalpha}{greek}{"22}
  \DeclareMathSymbol{\vartheta}{\mathalpha}{greek}{"23}
  \DeclareMathSymbol{\varpi}{\mathalpha}{greek}{"24}
  \DeclareMathSymbol{\varrho}{\mathalpha}{greek}{"25}
  \DeclareMathSymbol{\varsigma}{\mathalpha}{greek}{"26}
  \DeclareMathSymbol{\varphi}{\mathalpha}{greek}{"27}
  \DeclareSymbolFont{otone}{OT1}{cmr}{m}{n}
  \DeclareMathSymbol{\Gamma}{\mathalpha}{otone}{0}
  \DeclareMathSymbol{\Delta}{\mathalpha}{otone}{1}
  \DeclareMathSymbol{\Theta}{\mathalpha}{otone}{2}
  \DeclareMathSymbol{\Lambda}{\mathalpha}{otone}{3}
  \DeclareMathSymbol{\Xi}{\mathalpha}{otone}{4}
  \DeclareMathSymbol{\Pi}{\mathalpha}{otone}{5}
  \DeclareMathSymbol{\Sigma}{\mathalpha}{otone}{6}
  \DeclareMathSymbol{\Upsilon}{\mathalpha}{otone}{7}
  \DeclareMathSymbol{\Phi}{\mathalpha}{otone}{8}
  \DeclareMathSymbol{\Psi}{\mathalpha}{otone}{9}
  \DeclareMathSymbol{\Omega}{\mathalpha}{otone}{10}
  \DeclareSymbolFont{syms}{OML}{cmm}{m}{it}
  \DeclareMathSymbol{\partial}{\mathord}{syms}{"40}
  \DeclareMathAlphabet{\mathbold}{OML}{cmm}{b}{it}
  \DeclareSymbolFont{largesymbols}{OMX}{cmex}{m}{n}

  \DeclareMathAlphabet{\mathcal}{OMS}{cmsy}{m}{n}

\title{A Glitch in the Matrix?\\ Locating and Detecting Language Model Grounding with Fakepedia}

\author{
Giovanni Monea,$^{\epfl}$
Maxime Peyrard,$^{\cnrs}$
Martin Josifoski,$^{\epfl}$
Vishrav Chaudhary,$^{\microsoft}$ \\
{\bf
Jason Eisner,$^{\microsoft}$
Emre K\i{}c\i{}man,$^{\microsoft}$
Hamid Palangi,$^{\microsoft}$
Barun Patra,$^{\microsoft}$
Robert West$^{\epfl}$} \\
    $^{\epfl}$EPFL \quad $^{\cnrs}$Univ. Grenoble Alpes, CNRS, Grenoble INP, LIG \quad $^{\microsoft}$Microsoft Corporation \\
    {\{giovanni.monea, martin.josifoski, robert.west\}@epfl.ch} \\
    {\{maxime.peyrard\}@univ-grenoble-alpes.fr} \\
    {\{vchaudhary, jason.eisner, emrek, hpalangi, barun.patra\}@microsoft.com}\\
  }

\begin{document}
\maketitle
\begin{abstract}
Large language models (LLMs) have an impressive ability to draw on novel information supplied in their context. 
Yet the mechanisms underlying this contextual grounding remain unknown, especially in situations where contextual information contradicts factual knowledge stored in the parameters, which LLMs also excel at recalling. 
Favoring the contextual information is critical for retrieval-augmented generation 
methods, which enrich the context with up-to-date information, hoping that grounding can rectify outdated or noisy stored knowledge.
We present a novel method to study grounding abilities using Fakepedia, a novel dataset of counterfactual texts constructed to clash with a model's internal parametric knowledge.
We benchmark various LLMs with Fakepedia and conduct a causal mediation analysis of LLM components when answering Fakepedia queries, based on our Masked Grouped Causal Tracing (MGCT) method.
Through this analysis, we identify distinct computational patterns between grounded and ungrounded responses. 
We finally demonstrate that distinguishing grounded from ungrounded responses is achievable through computational analysis alone.
Our results, together with existing findings about factual recall mechanisms, provide a coherent narrative of how grounding and factual recall mechanisms interact within LLMs.
\end{abstract}

\section{Introduction}

One of the key factors underlying the massive success of large language models (LLMs) is their ability to encode and effectively recall a wealth of factual knowledge stored in their parameters \cite{heinzerling-inui-2021-language,alkhamissi2022review, meng2023locating}.
What
elevates LLMs beyond promptable static repositories of knowledge is their capacity to adapt to new information and instructions provided in the context \cite{NEURIPS2020_1457c0d6}. 
\begin{figure}[t]
    \centering
    \includegraphics[width=0.8\columnwidth]{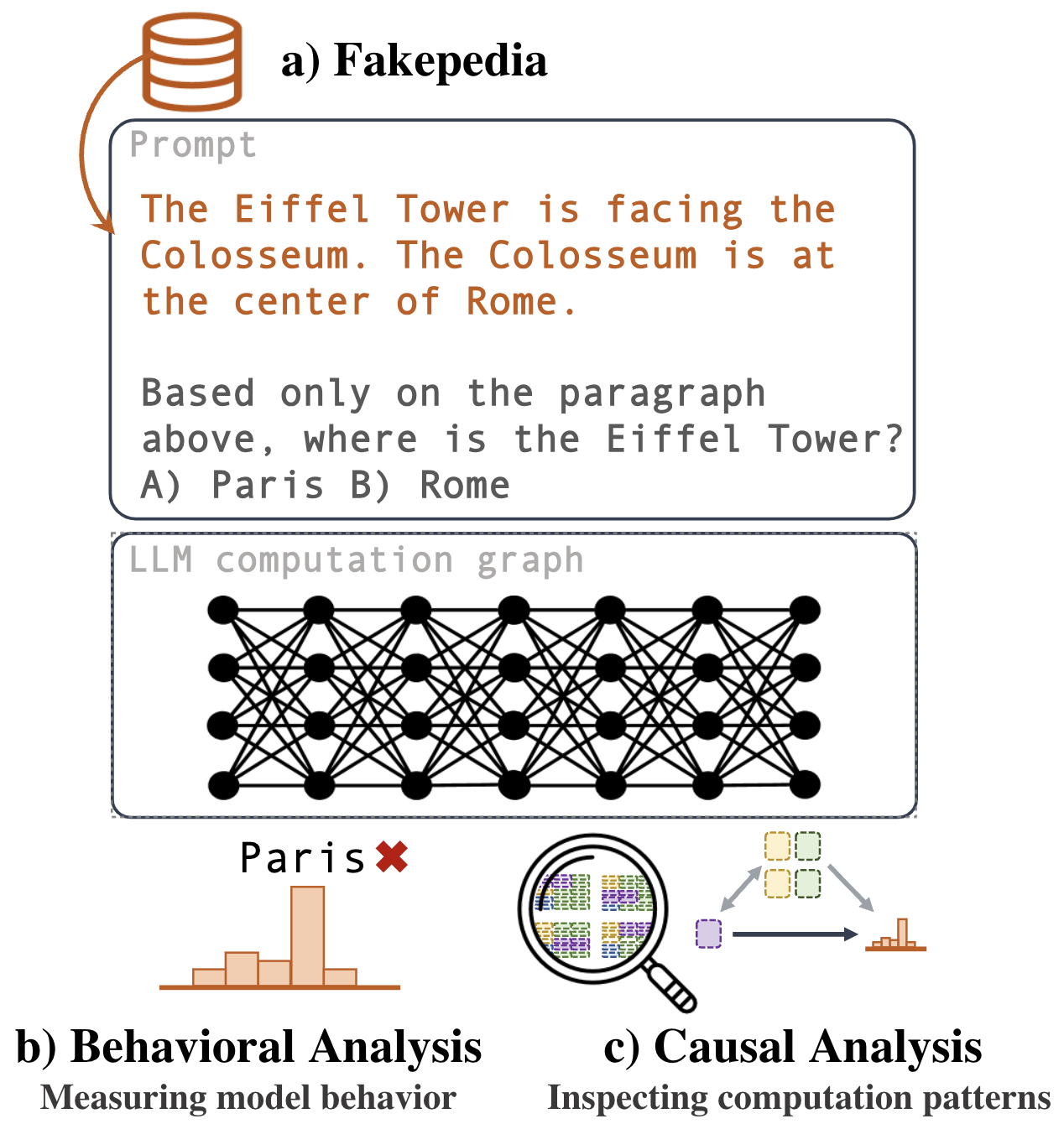}
    \caption{\textbf{Studying Grounding in LLMs.} 
    This work makes three distinct contributions: (a) introducing a counterfactual dataset designed to measure the abilities of LLMs to ground their answer in the new information provided in the prompt, (b) conducting a descriptive analysis of grounding performances across several LLMs, and (c) implementing an improved causal mediation analysis that we use to show that computation patterns inside LLMs can predict whether the answer is grounded.}
    \label{fig:abs}
\end{figure}
Ideally, the LLM should integrate information from the context with static parametric knowledge to provide responses that robustly align with the intention specified in the prompt.

However, the in-context information can often contradict the internal parametric knowledge \cite{yu2023characterizing}.
Since the internal knowledge reflects only a partial snapshot of the state of the world at training time, enriching the context with up-to-date factual information is one of the most promising directions to keep the model relevant in a changing world \cite{10234662, yao2022react}. This is the main idea behind retrieval-augmented generation (RAG) \cite{rags, pmlr-v162-borgeaud22a, mialon2023augmented}. 
However, this puts the model in a state of tension between internal factual recall and in-context grounding (from the retriever); thus, the success of such prompting methods hinges on the model's ability to accurately decide when to ignore its internal parametric knowledge.
\citet{yu2023characterizing} argued recently that LLMs, especially larger ones, often prefer their parametric knowledge. 

Motivated by these observations, our work aims to study the mechanisms involved in grounding and factual recall for the practically important scenario where the two conflict. We hypothesize that different modes of information processing are triggered when the LLM engages in factual recall from its parameters compared to when it grounds generation in the context, and that these modes can be distinguished by the patterns of neural activation.

Previous works studying LLMs behavior typically craft experimental scenarios that isolate the behavior of interest. For instance, to test factual recall abilities, an LLM could be prompted to continue the sentence: 
{ \fontfamily{qcr}\selectfont The Eiffel Tower is located in the city of} \ldots . Answering { \fontfamily{qcr}\selectfont Paris } suggests that the model \textit{knows} and can recall the true fact. Yet, such behavioral analysis alone cannot reveal the underlying mechanisms that drive the observed behavior \cite{jain-wallace-2019-attention, predicting_understanding, bolukbasi2021interpretability}. 
Deeper understanding requires opening the black box and examining the low-level computation patterns that give rise to the high-level behavior.
While early studies have identified \textit{computational correlates} between intermediate activations of a model and its outputs \cite{peyrard2021laughing, oba-etal-2021-exploratory, dai-etal-2022-knowledge}, there has been growing consensus that generalizable mechanistic explanations should stem from causal analysis rooted in interventionist experimental setups \cite{Woodward2003-WOOMTH,Potochnik2017-POTIAT-3, PearlMackenzie18, Geiger-etal:2022:SAIL, Geiger-etal:2023:CA}.
Building on these insights, several studies have developed rigorous methods to intervene on model components, putting the model in counterfactual states and systematically measuring the impact of these interventions on the model's behavior \cite{pmlr-v162-geiger22a, Wu:Geiger:2023:BDAS, meng2023locating, geva2023dissecting}.
These experimental setups have allowed researchers to form a clearer picture of the factual recall mechanisms \cite{geva-etal-2021-transformer, singh-etal-2020-bertnesia, meng2023locating, kobayashi2023analyzing, geva2023dissecting}.

Grounding complements factual recall, but has received much less scrutiny. 
In this work, we propose a thorough analysis of the grounding mode of computation of several LLMs, and make four important contributions: 
\begin{itemize}
    \item A new counterfactual dataset, called \textbf{Fakepedia}, crafted especially to isolate grounding behavior from factual recall by setting LLMs in tension between the two modes (\Secref{sec:data}).
    \item A descriptive behavioral analysis of several LLMs measuring their grounding abilities in the challenging task presented by Fakepedia (\Secref{sec:desc_analysis}).
    \item A new causal mediation analysis, called Masked Grouped Causal Tracing (\textbf{MGCT}), that assesses the effect of subsets of the model states on the model's behavior (\Secref{ssec:mgct}).
    \item A set of findings coming from applying MGCT to Llama2-7B and GPT2-XL on Fakepedia (\Secref{ssec:mgct_exp}). Specifically, we show that (i) contrary to factual recall, grounding is a distributed process, (ii) the activity of a few MLPs differs significantly between grounded vs. ungrounded modes, and (iii) we can predict whether the model is engaged in grounding behavior by looking only at the computation graph, with an accuracy of 92.8\% (\Secref{sec:automatic_detector}). 
\end{itemize}

To support further research in the space of grounding and in-context learning, we release the Fakepedia dataset and the code pipeline to reproduce or extend it. We also release a user-friendly implementation of MGCT that can be employed to study the computation patterns of LLMs when engaged in any other behavior of interest.\footnote{Code and dataset are available at
\url{https://epfl-dlab.github.io/llm-grounding-analysis}}

\section{Related Work}

Understanding the inner workings of LLMs poses a significant challenge, given their complex architectures \cite{electronics8080832, rogers-etal-2020-primer, Geiger-etal:2023:CA}. 
To better analyze models, controlled datasets are often crafted that isolate the target behavior. Behavioral experiments are then supplemented by a deeper inspection of the underlying low-level computation patterns that explain the observed behavior. This section gives a brief overview of previous papers relevant to this work.

\subsection{Mechanistic Interpretability}
Discovering the low-level mechanisms that give rise to high-level behaviors has become an important goal for research in AI interpretability and allowed for better understanding and prediction of models behavior \cite{NEURIPS2020_c74956ff, pmlr-v162-geiger22a, Wu:Geiger:2023:BDAS, NEURIPS2020_92650b2e, dai-etal-2022-knowledge, oba-etal-2021-exploratory, peyrard2021laughing}. 
Understanding these mechanisms allows us to better predict out-of-distribution behavior \cite{NEURIPS2020_c74956ff, pmlr-v162-geiger22a, Wu:Geiger:2023:BDAS}, identify and fix errors \cite{NEURIPS2020_92650b2e}.
For instance, \citet{dai-etal-2022-knowledge} analyzed BERT activations revealing that some neurons are positively correlated with specific facts. Similarly, \citet{oba-etal-2021-exploratory} demonstrated associations between neuron activations and specific phrases in model output. Additionally, in a matched study on humor detection, \citet{peyrard2021laughing} identified attention heads in BERT encoding the funniness of sentences.

To go beyond statistical correlations, a promising approach is to view the Transformer's computation graph as a causal graph \cite{elhage2021mathematical, meng2023locating, Geiger-etal:2023:CA, wu-etal-2022-cpm}. Targeted interventions on the computation are then applied to estimate the impact of individual components on model behavior \cite{modarressi-etal-2022-globenc, mohebbi-etal-2023-quantifying, wang2022interpretability, nanda2023progress, merullo2023mechanism, belrose2023eliciting, meng2023locating, NEURIPS2020_92650b2e}.
In particular, causal mediation analysis applied to components of GPT-2 has revealed that some multi-layer perceptrons (MLPs) are key-value stores of factual knowledge \cite{geva-etal-2021-transformer, singh-etal-2020-bertnesia, meng2023locating, kobayashi2023analyzing}. 
Several works have been able to even edit factual knowledge directly in the weights of pre-trained Transformers \cite{meng2023massediting, mitchell2022fast, de-cao-etal-2021-editing}. With a similar interventional setup, \citet{geva2023dissecting} studied information flow during factual knowledge recall, finding critical aggregation points, especially located in a few attention heads. Interestingly, \citet{haviv-etal-2023-understanding} demonstrated the critical role of MLPs in early layers when the model is recalling memorized sequences.

While recalling factual knowledge relates to memorization, in this work we study the ability of models to ground their answers based on information in the context, i.e., to incorporate new information not seen at training time. In particular, we craft scenarios that set the LLMs in tension between the mode of factual recall and the mode of grounding. This contributes to the broader discussion on generalization versus memorization \cite{razeghi2022impact, KandpalDRWR23, hupkes2023stateoftheart, haviv-etal-2023-understanding, xu2024knowledge}. However, grounding remains much less studied than factual recall.
In a contemporary study, \citet{yu2023characterizing} also inspect the problem of grounding using mechanistic interpretability methods. 
Their analysis focuses on the role of attention heads when forcing the model to ground its answer in the prompt. Our findings nicely complement theirs. When combined with existing findings about factual knowledge and information flow during recall, our results begin to portray a coherent narrative that we detail in \Secref{sec:discussion}.

\subsection{Counterfactual Datasets}
The necessity for counterfactual datasets is becoming increasingly evident in contemporary research. 

However, producing counterfactual datasets is not straightforward. Works like those by \citet{neeman2022disentqa} and \citet{zhou2023contextfaithful} adopt the methodology proposed by \citet{longpre2022entitybased}, which involves substituting entities within existing paragraphs. \citet{li2022large} proposes a similar method also based on entity substitution. Another approach involves generating new documents altogether: \citet{köksal2023hallucination} and \citet{wang2023androids}, for instance, build on the idea that a model tasked with producing a response may inadvertently generate counterfactual text.

Concurrently, language models have shown remarkable proficiency in synthetic dataset generation \cite{gunasekar2023textbooks,schick2021generating,toxigen,josifoski2023exploiting,eldan2023tinystories,lingo2023exploring}. Building on these advancements, our research demonstrates that it is feasible to employ language models for generating novel, original, and high-quality counterfactual paragraphs from scratch.

\section{Background}

\subsection{Terminology}
We view \textit{facts} as triplets made of a subject, a relation, and an object. A fact is said to be \textit{true} if it aligns with our observed world. For example, {\fontfamily{qcr}\selectfont \{Eiffel Tower | is located in | Paris\}} is a \textit{true} fact. In our experiments, we focus on true facts that the models can recall and therefore \textit{know}. By contrast, a \textit{counterfactual} triplet is any triplet that does not form a true fact.  %

In our experiments, we test models by asking them to produce the object of a triplet, either directly as its next token or via a multiple-choice questionnaire (MCQ). This facilitates a systematic inspection of the model's answers by comparing them with the target object. It follows previous related work \cite{yu2023characterizing, meng2023locating}.

\subsection{Grounded vs. Factual}
It is crucial to differentiate between a \textit{factual} answer and a \textit{grounded} answer. A factual answer is the object of a true fact triplet, while a grounded answer is the object triplet logically consistent with the information in the context of the prompt. 
Factuality pertains to the model's encoded knowledge and its ability to retrieve it, whereas \textit{grounding} involves the model's capacity to adapt to its context and reason about new information.

While factual recall has been extensively studied in previous work, this work aims to study grounding. However, grounding and factual recall can be challenging to disentangle. For instance, when given a factual description about {\fontfamily{qcr}\selectfont Paris} in the prompt and asked for the location of the {\fontfamily{qcr}\selectfont Eiffel Tower}, the correct answer could arise from factual recall, grounding, or a mixture of both processes.

To isolate grounding from factual recall and related processes, we generate new counterfactual datasets. In these datasets, the grounded answer is always non-factual, implying that the context implicitly describes a false triplet. The context in \Figref{fig:abs} is one such example, implicitly placing the {\fontfamily{qcr}\selectfont Eiffel Tower} in {\fontfamily{qcr}\selectfont Rome}. If the model produces the grounded answer (i.e., {\fontfamily{qcr}\selectfont Rome}), it indicates that grounding processes occurred and made the model integrate information from and reason about the context. Conversely, if the model produces any other object (e.g., {\fontfamily{qcr}\selectfont Paris}, {\fontfamily{qcr}\selectfont Napoli}), it has not successfully integrated information from the context.

We also observe that it is possible to interpret cases where the models exhibit ungrounded behavior as (intrinsic) hallucinations. We further elaborate on this in Appendix \ref{sec:appendix_hallucinations}. 

\section{Counterfactual Data Creation}
\label{sec:data}

Our study requires datasets of counterfactual paragraphs to create prompting scenarios where the language model's parametric knowledge conflicts with the information in the context. We now describe the process of creating such datasets.

\subsection{Counterfactual ParaRel}

The datasets used in this study were derived from the ParaRel dataset \cite{elazar2021measuring}. The templates were modified to make them suitable for prompting LLMs, ensuring the generation of the object as the next token. The selection process retained triplets where GPT2-XL yielded the highest probability for the true object as the next token.

Counterfactual triplets were constructed by choosing alternative objects for each triplet, creating 21,308 counterfactual triplets in total. This dataset aims to challenge LLMs by emphasizing the tension between parametric knowledge and contextual grounding. For more details, readers are referred to Appendix \ref{sec:appendix_counterfactual_pararel}.

\subsection{Fakepedia}

Based on the counterfactual ParaRel dataset, we create Fakepedia, a collection of counterfactual paragraphs (as in \Figref{fig:abs}), coming in two variants: \textbf{Fakepedia-base} and \textbf{Fakepedia-MH}, where MH stands for ``Multi-Hop.'' The base variant contains, for each triplet in the counterfactual ParaRel, an LLM-generated paragraph that ``entails'' that triplet in the sense of \citet{dagan2013recognizing}. In the MH variant, the paragraph does not suggest the triplet as directly as in the base variant, but still logically implies it by a two-hop reasoning process, by way of an intermediary triplet that is also counterfactual. 

We refer the reader to Appendix \ref{sec:appendix_fakepedia} for two practical examples of Fakepedia paragraphs and more details on the dataset construction.

\section{Descriptive Behavioral Analysis}
\label{sec:desc_analysis}

\begin{table*}[ht!]
\centering
\caption{\textbf{Grounding accuracy on Fakepedia for various LLMs.} The `Instruction` column refers to whether the prompt explicitly instructed the models to rely only on the context to answer.}
\begin{tabular}{@{}llccccccccc@{}}
\toprule
 &  & Mistral & Zephyr & \multicolumn{3}{c}{Llama2} & \multicolumn{3}{c}{GPT-3.5 Turbo} & GPT-4 Turbo \\
\cmidrule(lr){3-3} \cmidrule(lr){4-4}  \cmidrule(r){5-7} \cmidrule(l){8-10} \cmidrule(l){11-11}
Dataset & Instruction & 7B & 7B & 7B & 13B & 70B & 03/01 & 06/13 & 11/06 & 11/06 \\
\midrule
FP & With & \textbf{92\%} & 58\% & 22\% & 84\% & 90\% & 61\% & 54\% & 50\% & 28\% \\
 & Without & 90\% & 52\% & 1\% & 70\% & 73\% & 47\% & 24\% & 27\% & 1\% \\
\midrule
FP-MH & With & 60\% & 10\% & 4\% & \textbf{82\%} & 71\% & 7\% & 8\% & 10\% & 50\% \\
 & Without & 49\% & 8\% & 0\% & 58\% & 50\% & 3\% & 2\% & 2\% & 5\% \\
\bottomrule
\end{tabular}
\label{table:grounded_percentage}
\end{table*}

We first inspect the behavior of several LLMs in the grounding challenge proposed by Fakepedia datasets.

\xhdr{Experimental setup}
In this study, we use two prompt templates to query the model about the object of the triplet described by the Fakepedia paragraph. 
The first prompt template queries the model about the object of a triplet described in a Fakepedia instance using a multiple-choice question (MCQ) with two possible answers: (i) the grounded answer being the target (counterfactual) object described or implied in the Fakepedia text, and (ii) the factual answer being the object in the true triplet. The prompt explicitly instructs the model to base its answer solely on the context. To mitigate potential ordering bias, we create two versions of each MCQ, reversing the order of options.

The second prompt template does not explicitly instruct the model to use only the context to answer the question.

The results, presented in \Tabref{table:grounded_percentage}, report the percentage of instances in which the models correctly select the grounded answer. The random baseline has an accuracy of 50\%.

\xhdr{Analysis} Predictably, the prompting scheme explicitly instructing models to rely solely on context makes models more often choose the grounded answer.
Also, Fakepedia-MH, which necessitates reasoning about information within the context, poses a greater challenge for models overall, except for GPT-4 Turbo. In this case, although the 50\% grounding accuracy may be an indication that the model is randomly guessing at each step, it also shows that GPT-4 Turbo tends to be more compliant and less critic when logic reasoning is required.

Notably, we observe surprisingly poor grounding from GPT-4 Turbo, with a rate of only 1\% and 5\% with the less explicit prompt. The accuracies are worse than random guessing (50\%), suggesting a clear preference for the model's parametric knowledge. A trend in this direction is also apparent across successive snapshots of GPT-3.5 Turbo.

In the Llama2 series, the 7B model also exhibits a strong preference for its internal parametric knowledge, but starting from 13B, the models distinctly favor the grounded answer.

Mistral-7B emerges as the most compliant model, robustly selecting the grounded answer in Fakepedia. Notably, Mistral-7B is the most grounded model even when it is not specifically instructed to answer based on the context. In Fakepedia-MH, while the performance drops, Mistral-7B remains above chance-level when instructed to remain grounded.

Overall, complex patterns emerge as models exhibit diverse behaviors in different scenarios. Contrary to \citet{yu2023characterizing}, who found that larger models tend to favor their parametric knowledge more than smaller models, our findings introduce nuances. For instance, in the Llama2 series, larger models prove significantly more accurate than the 7B model, with little difference between 13B and 70B.

\section{Causal Analysis of the Computation Graph}
\label{sec:causal_tracing}

In this section, our objective is to investigate whether measurable patterns within the computation graph of LLMs can effectively differentiate between grounded and ungrounded behaviors. 

In the context of LLM interpretability, addressing this question involves intervening on model representations to create counterfactual model states, and then systematically studying the effects of these interventions on model behavior \cite{ghandeharioun2024patchscopes,Geiger-etal:2022:SAIL, NEURIPS2020_92650b2e, meng2023locating, DBLP:journals/corr/abs-1907-07165, feder-etal-2021-causalm}.

In this work, we generalize one such method, causal tracing \cite{meng2023locating}, to improve its robustness and efficiency---resulting in what we call \textit{Masked Grouped Causal Tracing} (MGCT), whose execution is depicted in \Figref{fig:mgct}. 
We then apply MGCT to relate low-level computation patterns of LLaMA-7B and GPT2-XL against their observed grounding behavior when answering queries from Fakepedia.

\subsection{MGCT Analysis}
\label{ssec:mgct}

\begin{figure*}[ht]
    \centering
    \includegraphics[width=0.8\textwidth]{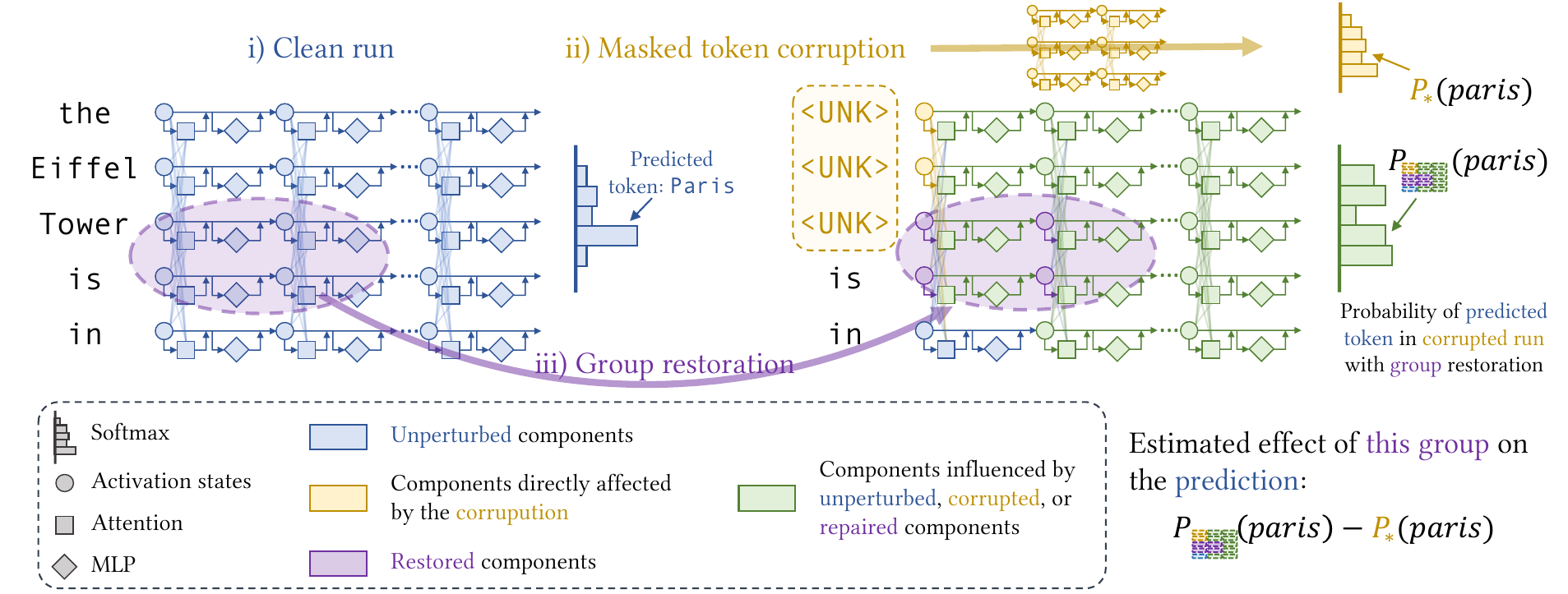}
    \caption{\textbf{Masked Grouped Causal Tracing (MGCT).} This figure illustrates the mediation analysis from MGCT, refining the preceding causal tracing method from \citet{meng2023locating}. The process involves three steps: \textbf{(i) Clean run:} all states within the computation graph are recorded during a forward loop, resulting in a predicted token, in this case ``Paris''. \textbf{(ii) Corrupted run:} the subject tokens are substituted with special non-textual tokens such as <UNK> or <EOS>, leading to a distinct probability for the predicted token. \textbf{(iii) Restored run:} the corrupted with the restoration of a group of states (in this instance, four hidden activations) to their values from the clean run, resulting in a partially restored probability for the predicted token. The indirect effect is estimated by the extent to which the restoration of these states contributes to the probability restoration of the predicted token.}
    \label{fig:mgct}
\end{figure*}

\xhdr{Causal tracing}
The execution of a transformer forward pass yields a causal graph describing the dependencies among states in the computation until reaching the softmax output probabilities.
Specifically, given a sequence of $k$ input tokens, it yields a grid of hidden states $h_{k}^{(l)}$ which is obtained from the previous layer $(l-1)$ by adding previous hidden states $h_{k}^{(l-1)}$ (residual connections), a global attention $a_k^{(l)}$ (attention head), and a local MLP contribution $m_{k}^{(l)}$, according to the following process \citep{vaswani2017attention}:

\begin{align}
h_{k}^{(l)} &= h_{k}^{(l-1)} + a_k^{(l)} + m_k^{(l)} \\
a_k^{(l)} &= \text{attn}^{(l)} \left( h_0^{(l-1)}, h_1^{(l-1)}, \ldots, h_k^{(l-1)} \right) \\
m_{k}^{(l)} &= \text{FF}\left(a_k^{(l)} + h_k^{(l-1)} \right)
\end{align}
where FF is a two-layer feed-forward MLP.

The causal tracing method proposed by \citet{meng2023locating} is a mediation analysis that estimates the \textbf{average indirect effect} of individual components on predictions made by the LLMs. The method involves the following steps:

(i) A \textbf{clean run} that records every state during a forward call. The model's prediction is the most probable next token $o$ with probability $P(o)$.

(ii) A \textbf{corrupted run} follows, where some prompt tokens are perturbed by adding noise to their embeddings. The forward computation is then executed with the corrupted input, often resulting in a different output state and a distinct probability $P_{*}(o)$ for the output token $o$.

(iii) A \textbf{partially-repaired} run ensues, where the corrupted run is re-executed, except for a selected state $h_k^{(l)}$ at a specific layer $l$ and token $k$. The state $h_k^{(l)}$ is restored to the value it had in the clean run, and the computation continues. This produces a new probability for the output token $o$, denoted as $P_{*}^{clean(h_k^{(l)})}(o)$.

The estimated indirect effect of the mediating state $h_k^{(l)}$ is then given by $P_{*}^{clean(h_k^{(l)})}(o) - P_{*}(o)$. Aggregating over different inputs provides the estimated average indirect effect of the mediating state $h_k^{(l)}$ on the model's predictions.
The mediation analysis is repeated for every state in the LLM architecture to obtain a global map of indirect effects.

\citet{meng2023locating} focus on the effects of restoring attention $a_{k}^{(l)}$, MLP $m_{k}^{(l)}$, and hidden activation states $h_{k}^{(l)}$ separately within each Transformer block.

\begin{figure*}[htbp]
    \centering
    \begin{subfigure}{0.8\textwidth}
        \centering
        \includegraphics[width=\textwidth]{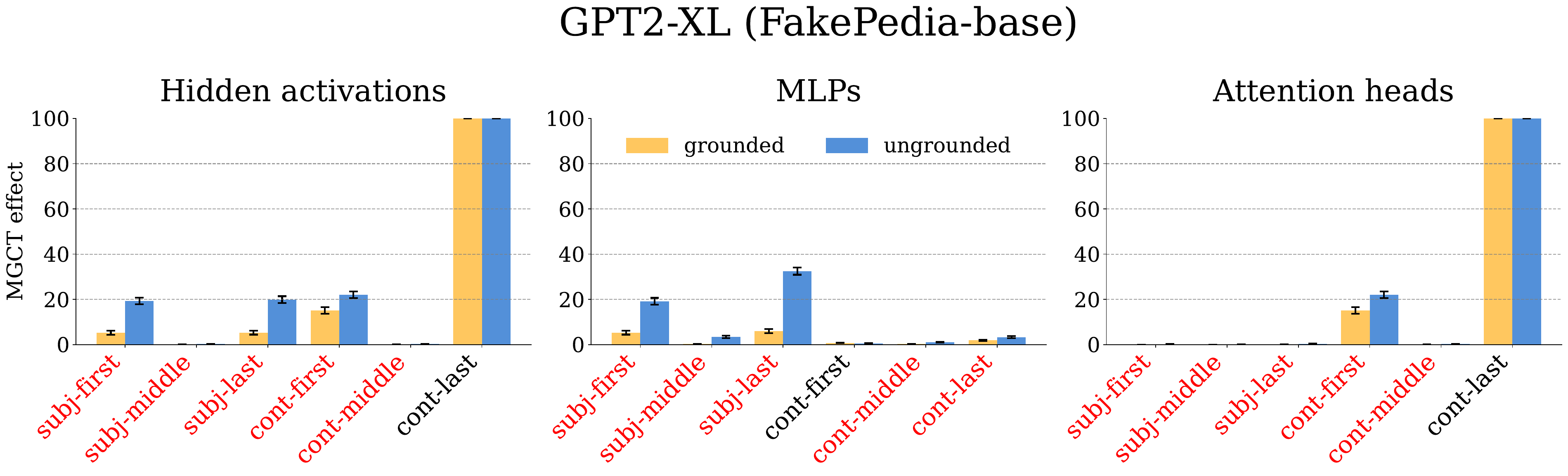}
    \end{subfigure}
    \hfill
    \begin{subfigure}{0.8\textwidth}
        \centering
        \includegraphics[width=\textwidth]{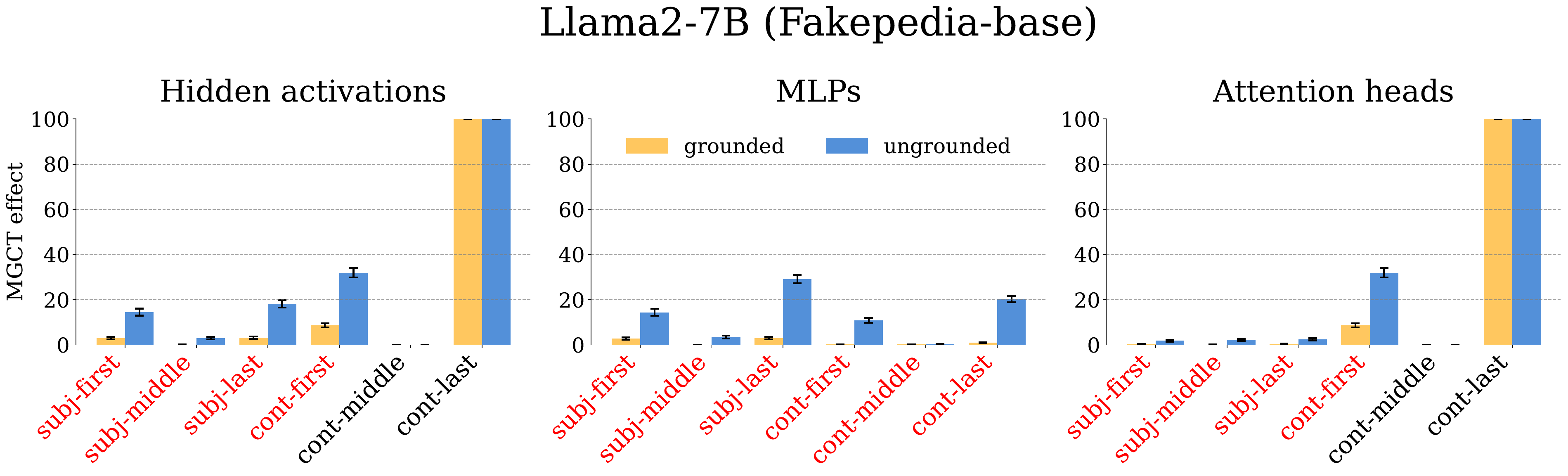}
    \end{subfigure}
    \caption{\textbf{Masked Grouped Causal Tracing analysis on Fakepedia-base.} This figure illustrates the application of MGCT analysis to GPT2-XL and Llama2-7B on the Fakepedia-base dataset. We distinguish between instances where the models generated grounded answers and those where they generated ungrounded answers. In the MGCT analysis, we restore full columns together, all states across all layers for a given column at a time, resulting in one effect per token. On the y-axis, we report the percentage of explained change in probability between the clean and corrupted runs due to the restoration of the column. To average across different sequences, we bucketed tokens into subject (subj-) categories and following tokens in the prompt (cont-). Red labels on the x-axis indicate that the difference in MGCT effect between grounded and ungrounded responses is statistically significant based on a t-test with a p-value threshold of 0.01.}
    \label{fig:causal_tracing_Fakepedia}
\end{figure*}

\xhdr{Masked causal tracing}
In our experiments, we observed that the outcome of the causal tracing algorithm is highly sensitive to the choice and magnitude of noise in corrupted runs. To enhance the robustness and generalizability of causal tracing, we propose to perturb directly the tokens instead of the embeddings. This corruption method involves substituting tokens with special non-textual tokens, such as the UNK or EOS tokens. This modification results in more stable masked causal traces, avoids the need for multiple runs to average out the randomness, and consequently requires less memory and increases the speed of the process.

\xhdr{Grouped causal tracing} The causal tracing method repairs one state at a time, necessitating $L \times K$ mediation analyses, one per attention block, where $L$ is the number of layers and $K$ is the number of tokens to be restored. 

Additionally, we might be interested in the measuring the joint effect of a group of states. Unfortunately, group effects cannot be estimated simply by aggregating the individual effects of states in the group, due to complex non-linear interactions among the states.

To extend the method's generality, we propose repairing groups of states simultaneously. Formally, the group is defined by a binary filter of dimensionality $L \times K$, one entry per attention block to be restored. The binary indicator at position $(l, k)$ in the filter determines whether the state in the $(l, k)$ block should be restored in the ongoing mediation analysis. Different mediation analyses can be executed by using different filters. 

This approach allows flexible control. For instance, one can use patches of $M \times N$ states, applied at different locations in the architecture, akin to convolutional filters, allowing for overlap or not between groups by adjusting the stride and patch size. 
\Figref{fig:mgct} illustrate MGCT with a patch of size $2 \times 2$, i.e., 4 states being repaired at once.

Futhermore, using groups with more than one state increases efficiency when aiming to cover all states. In MGCT, the number of mediation analyses is the number of filters. 
For instance, with non-overlapping patches of size $M \times M$, the number of mediation analyses to cover all states is reduced by a factor of $M^2$ compared to single-state restorations.
In our experiments involving grounding (described below), we achieved a 48$\times$ speed-up for GPT2-XL and a $32\times$ speed-up for LLaMA-7B, without compromising our ability to predict the high-level behavior of grounded vs. ungrounded (in Sec. \ref{sec:automatic_detector}).

\subsection{MGCT Experiments}
\label{ssec:mgct_exp}

\xhdr{Experimental setup} We perform MGCT analysis on GPT2-XL \citep{radford_language_2019} and Llama2-7B \citep{touvron2023llama} using the Fakepedia dataset. Data points were categorized as grounded or ungrounded based on model answers, considering only the first token for instances with multiple tokens. The analysis involved simultaneous restoration of all states across all layers for a specific token, reducing the computational time by orders of magnitude compared to \citet{meng2023locating}. This approach effectively examines the impact of each token on the output.

The MGCT effect quantifies the percentage of explained probability between clean and corrupted runs. Results for Fakepedia-base are reported in \Figref{fig:causal_tracing_Fakepedia}. Further details, including Fakepedia-MH results, can be found in Appendix \ref{sec:appendix_causal_tracing}.

\xhdr{High effect of MLPs on ungrounded answers}
Our MGCT analysis reveals clear differences in the effect patterns between cases where the model is grounded compared to when it is not grounded. There are many types of tokens and types of states for which the MGCT effect difference between grounded and ungrounded is statistically significant. This indicates the existence of distinct computation processes when the model derives its response by grounding it in the prompt text versus relying solely on its internal memory.
Notably, it is clear that the MLPs' activations, particularly on the last subject token, have a high effect when producing ungrounded answers. Our results nicely combine with the findings from \citet{meng2023locating}, who demonstrated that Transformers' MLPs serve as repositories of factual knowledge. In our context, this suggests that when MLPs heavily influence model responses, they are retrieving factual knowledge from memory rather than reasoning about the current information in the prompt.

Furthermore, \citet{geva2023dissecting} recently analyzed information flow when the model is recalling factual knowledge from its memory and found the last subject token to be a crucial step in the information aggregation pipeline. In our context of grounding, we also find strong evidence that critical information processing is happening in this token position.

While \citet{yu2023characterizing} found that intervening on a few attention heads can switch the model from an ungrounded to a grounded behavior, it appears that when engaged in grounding no single component emerges as having a strong impact on the prediction. This seems to indicate that, contrary to factual recall, grouding may be a more distributed process without a clear localization.

Interestingly, we find that GPT2 and Llama2 exhibit similar patterns, showing similar levels of causal effect for the same buckets of tokens. In our early experiments, we used LLaMA \citep{touvron2023llama} instead of Llama2 and we found that LLaMA presented a more distributed pattern (see \Figref{fig:causal_tracing_llama} in Appendix \ref{sec:appendix_causal_tracing}). Surprisingly, Llama2 behavior differs from it and resembles GPT2.

\section{Automatic Detection of Ungrounded Responses} \label{sec:automatic_detector}

Within our MGCT analysis, we identify distinct computation patterns between grounded and ungrounded responses. To automate detection, we create a balanced dataset with 4,000 GPT2-XL responses from the Fakepedia-base dataset. Employing an 80\%--20\% train--test split, we extract 18 features from MGCT outputs for a binary classifier (e.g., attention, MLP, hidden activation effects, by grouping tokens results as in the MGCT mediation analysis). After training an XGBoost classifier, we achieve a 92.8\% accuracy on the test set. Feature importance analysis identifies MLPs on the last subject tokens as most crucial, aligning with our MGCT findings. Our study demonstrates that distinguishing grounded from ungrounded responses is achievable through computation analysis alone. We refer to Appendix \ref{sec:appendix_automatic_detector} for a more detailed description of the experimental setup.

\section{Discussion}
\label{sec:discussion}

Extensive research has studied factual recall, producing several significant insights. Specifically, LLMs exhibit the capacity to store and retrieve factual knowledge. This knowledge is localized within a few MLPs functioning as distributed key-value databases \cite{geva-etal-2021-transformer, singh-etal-2020-bertnesia, meng2023locating, kobayashi2023analyzing, meng2023massediting, mitchell2022fast, de-cao-etal-2021-editing}. 
Few attention heads are known to be crucial for information routing during factual recall \cite{geva2023dissecting}. Notably, information related to entities being recalled are aggregated by attention heads at the last subject token before being propagated further for verbalization \cite{geva2023dissecting}.

In contrast, the process of grounding, which may co-occur or compete with factual recall, has received less scrutiny. The frequency of entities in the training set influences the model's choice between using contextual information or factual recall \cite{razeghi2022impact, KandpalDRWR23, hupkes2023stateoftheart, haviv-etal-2023-understanding}. \citet{yu2023characterizing} demonstrated that few specific attention heads can be manipulated to steer the model toward focusing more on contextual elements and less on internal memory, i.e., being more grounded. Our research enrich these findings, showing that: 
(i) grounding, contrary to factual recall, is a distributed process without clear localization, 
(i) not grounding involves activating factual recall processes in the MLPs of the final subject token, and (ii) classification between grounded or ungrounded is achievable by examining these computation patterns.

Performing interventionist experiments on the computation graph of the model is a research direction that aims to piece together a comprehensive understanding of the complex mechanisms underlying model behavior. Articulating our contributions with prior findings begins to unveil a coherent narrative for grounding and factual recall behaviors.

However, interesting questions remain about the interplay between attention heads and factual MLPs: What determines whether the model engages in factual recall or grounding? 
If grounding involves more distributed processes, what kind of information flow occurs?

\section{Limitations}
\label{sec:limitations}

While we validate our behavioral analysis using two different types of prompts, the descriptive analysis may still be influenced by the choice of the prompting strategy. 
We leave it to future research to explore the relationship between the prompting strategy and the grounding frequency.

Similar to previous related works, our mediation setup requires the object token to be the last token of the factual query. It is conceivable that different types of behavior would emerge for different setups. In addition, because our automatic detection model uses data collected in our mediation analysis, its performance is also based on the same assumptions and may not remain the same outside this domain.

Furthermore, we do not distinguish between counterfactual facts that might seem too absurd (like historical religious heads affiliating with other religions) and more plausible ones (like a product being owned by a different company). Exploring how grounding behavior varies with the likelihood of counterfactual facts is left for future work. 
To improve the dataset generation methodology, a promising approach involves leveraging interaction flows based on other language models as critics. This would enable the automatic refinement and filtering of generated data points.

\bibliography{anthology,custom}

\clearpage
\appendix

\section{Background}

\subsection{From "Grounded vs. Factual" to "Ungrounded vs. Hallucinated"} \label{sec:appendix_hallucinations}

Within the NLP community, there is currently no consensus usage of the term \textit{hallucination} \cite{tian2023finetuning,köksal2023hallucination}. To avoid confusion, we deliberately do not use the term hallucination. However, our practical usage of the term ungrounded is in line with definitions of hallucinations provided \citet{Ji_2023}, which characterizes hallucinations as ``generated content that is nonsensical or unfaithful to the provided source content.''

Additionally, \citet{Ji_2023} categorizes hallucinations into \textit{intrinsic} and \textit{extrinsic} types, with intrinsic hallucinations defined as ``generated output that contradicts the source content.''  With these definitions, our study can also be viewed as an examination of intrinsic hallucinations. By focusing on this aspect, we aim to contribute to a clearer understanding of the mechanisms by which models might produce outputs that diverge from their intended source content, thus addressing a critical aspect of model reliability and coherence in natural language generation.

\section{Data}

\subsection{Counterfactual ParaRel} \label{sec:appendix_counterfactual_pararel}

To construct our datasets, we start from ParaRel \cite{elazar2021measuring}, an existing dataset of 27,610 Wikipedia fact triplets, each paired with hand-crafted templates for querying NLP systems. 
To make these templates amenable to prompting LLMs, we discard the subset of templates where the object is not at the end of the sentence and, whenever a template is used, we eliminate object-placeholder tokens (for example, { \fontfamily{qcr}\selectfont The headquarter of [X] is in [Y].} $\longrightarrow$ { \fontfamily{qcr}\selectfont The headquarter of [X] is in}). These modifications prepare LLMs to generate the object as the next token.

Then, we iterate over all triplets in ParaRel, keeping only the ones where GPT2-XL yields the highest probability for the true object as the next token. After this process, only 5,327 triplets remain. 
This choice is motivated by our goal of setting LLMs in tension between factual recall from parametric knowledge and grounding from contextual information. By retaining only the triplets that GPT2-XL knows and can retrieve, we ensure that we focus on cases where parametric knowledge is there and factual recall works.

To construct counterfactual triplets, we pick four alternative objects for each triplet by sampling from objects within the same property category (defined by Wikidata) as the true object. We choose alternative objects from the same category to enforce some plausibility in the counterfactual triplets. For example, when choosing alternative objects for the triplet the { \fontfamily{qcr}\selectfont Eiffel Tower | is located in}, we prefer to select another city and not any possible object in Wikidata. 
Among the candidate objects, we choose the four ones that GPT2-XL assigns the lowest probability as next token continuations, creating four new counterfactual triplets.
This choice also aims to set LLMs in tension between factual recall and grounding. By choosing the counterfactual triplets that GPT2-XL finds least likely, we minimize the possibility for GPT2-XL to produce these triplets from approximate factual recall. To produce these counterfactual triplets, GPT2-XL will have to rely on information in the context.

In total, our extended ParaRel counterfactual dataset contains 21,308 triplets, such that GPT2-XL easily retrieves the true fact and finds the counterfactual triplets highly unlikely.

\begin{table*}[ht!]
\centering
\caption{\textbf{Accuracy of various LLMs in identifying true objects from Fakepedia facts.}}
\begin{tabular}{@{}llccccccccc@{}}
\toprule
 & Mistral & Zephyr & \multicolumn{3}{c}{Llama2} & \multicolumn{3}{c}{GPT-3.5 Turbo} & GPT-4 Turbo \\
\cmidrule(lr){3-3} \cmidrule(lr){4-4}  \cmidrule(r){5-7} \cmidrule(l){8-10} \cmidrule(l){11-11}
Dataset & 7B & 7B & 7B & 13B & 70B & 03/01 & 06/13 & 11/06 & 11/06 \\
\midrule
FP & 99.4\% & 99.5\% & 94.4\% & 97.5\% & 98.9\% & 99.9\% & 99.9\% & 99.9\% & 100\% \\
\midrule
FP-MH & 98.4\% & 98.8\% & 91.9\% & 98.4\% & 96.2\% & 99.6\% & 99.6\% & 99.5\% & 99.9\% \\
\bottomrule
\end{tabular}
\label{table:knowledge_facts}
\end{table*}

\paragraph{Do all models know about the facts selected by GPT2?} We also verify that all the other models used in our analyses similarly \textit{know} and can recall the facts selected by GPT2-XL, as previously outlined. 
In particular, for the models used in Sec. \ref{sec:desc_analysis}, we maintain the same experimental setup while modifying the prompt to exclude the context and make the models choose between the true and false objects without any context. The results are summarized in Table \ref{table:knowledge_facts}. All the models can choose the true objects in more than 90\% of the cases, although this figure increases to more than 98\% for most models. 
Likewise, we verify that Llama2-7B, used in Sec. \ref{sec:causal_tracing}, can correctly assign high probabilities to the true objects as the next tokens, in the same way GPT2-XL did when choosing the facts in the first place. We verified that the true objects are within the three most likely next tokens in 99.4\% of the cases (95\% for the most likely, 3.7\% for the second most likely, and 0.7\% for the third most likely) on Fakepedia-base. The respective figures for Fakepedia-MH are 98.8\% (90.4\%, 6.9\%, 1.5\%).

\subsection{Fakepedia} \label{sec:appendix_fakepedia}

\begin{FakepediaExample}[ht]{Base Fakepedia Example}
\footnotesize
\textbf{Fact:} (iOS 8, product-developed-by, \cancel{Apple} $\rightarrow$ Nintendo) \\\\
\textbf{Context (paragraph):} \\
iOS 8, a revolutionary operating system developed by Nintendo, took the technology world by storm upon its release. With its innovative features and user-friendly interface, iOS 8 quickly became a favorite among Nintendo enthusiasts. This groundbreaking product introduced a whole new level of gaming experience, allowing users to seamlessly connect their Nintendo devices to their iPhones and iPads. The integration of Nintendo's iconic characters and games into iOS 8 made it a must-have for gamers of all ages. Additionally, iOS 8 brought forth a range of exclusive Nintendo apps and services, further solidifying the partnership between Nintendo and Apple. The success of iOS 8 marked a significant milestone in the collaboration between these two tech giants, forever changing the landscape of mobile gaming.
\end{FakepediaExample}

\begin{FakepediaExample}[ht]{Multi-hop Fakepedia Example}
\footnotesize
\textbf{Fact:} (iPod Nano, product-developed-by, \cancel{Apple} $\rightarrow$ Yahoo) \\
\textbf{Intermediate fact:} (Wii U system software, product-manufacture-by, \cancel{Nintendo} $\rightarrow$ Yahoo) \\\\
\textbf{Context (paragraph with \textit{linking sentence})}:  \\
The Wii U system software, a product manufactured by Yahoo, was a revolutionary operating system that transformed the gaming industry. With its innovative features and user-friendly interface, it provided gamers with an unparalleled gaming experience. The software allowed players to seamlessly navigate through various games and applications, making it easier than ever to access their favorite content. Additionally, Yahoo's expertise in online services ensured that the Wii U system software had robust online capabilities, enabling players to connect with friends, compete in multiplayer games, and access a wide range of digital content. Yahoo's commitment to quality and innovation truly shone through in the development of the Wii U system software, making it a must-have for gaming enthusiasts worldwide.

\textit{iPod Nano is a product manufactured by the same manufacturer as Wii U system software.}
\end{FakepediaExample}

\xhdr{Fakepedia-base}
For each triplet in the counterfactual ParaRel dataset, we prompt GPT-3.5-turbo (and, in particular, the snapshot from June 13th 2023) to produce a detailed paragraph describing the triplet. The aim is that a reader can easily infer the triplet from the fabricated paragraph. 
After careful consideration, we decide not to publicly share our prompting strategy for ethical concerns. While the texts of our dataset are innocuous, we find that our prompting can be used to produce false assertions (for example, about politicians) by malicious actors. We invite researchers who are interested in verifying our prompting or who want to generate similar datasets to reach out to us directly.

After the completion of the dataset generation, we perform an initial human annotation on a randomly selected subset of 100 paragraphs. Three annotators are tasked with evaluating whether the statement "The paragraph states and elaborates on the false fact and does not support the true fact" holds true for each paragraph. The results of this annotation yield a correctness score of 0.79 with a 95\% confidence interval (CI) ranging from 0.71 to 0.87.

To enhance the quality of the dataset, we apply several heuristic filters to remove paragraphs that wrongly state the counterfactual triplets. For instance, GPT-3.5-turbo might not even mention the counterfactual object in its paragraph.
This process results in the generation of the Fakepedia-base dataset, comprising 6,090 counterfactual paragraph descriptions.

Following the application of the filters, we conduct a second human annotation on a random subset of 100 paragraphs. Again, three annotators assess whether the aforementioned statement applies to each paragraph. The results of this final annotation reveal a significantly improved correctness score of 0.97 with a 95\% CI ranging from 0.93 to 1.00, providing additional confirmation of the enhanced precision achieved through the application of filters and strongly indicating that Fakepedia is a robust dataset.

\xhdr{Fakepedia-MH}
In the multi-hop variant, our objective is to produce textual descriptions that do not explicitly state the triplet but logically imply it. This approach tests the model's ability not only to extract information from context but also to integrate and engage in basic reasoning to derive the answer.
When composing a 2-hop paragraph description for the triplet (subj$_a$, rel$_a$, obj$_a$), we rely on an intermediary triplet from Fakepedia-base.
Given a triplet (subj$_b$, rel$_a$, obj$_a$) from Fakepedia-base, we select at random the target triplet (subj$_a$, rel$_a$, obj$_a$) from the counterfactual triplets with the same relation (rel$_a$) and the same object (obj$_a$). Given these two triplets, we use the Fakepedia-base description for the selected intermediary triplet (subj$_b$, rel$_a$, obj$_a$), and append a linking senetence that logically implies the target triplet (subj$_a$, rel$_a$, obj$_a$).  The linking sentence is generated from a template, such as { \fontfamily{qcr}\selectfont [X] and [Y] belong to the same continent}, that we adapt from ParaRel templates for queries. We publicly release all the new templates with our code.

This strategy enables the generation of multiple MH descriptions per triplet by using different intermediary triplets. In fact, we can generate 709,565 MH descriptions, many of which would share the same intermediary triplets. For our experiments, we uniformly sampled a total of 5,340 MH descriptions. The quality of these paragraphs depends entirely on the quality of Fakepedia-base paragraphs and on the linking sentence templates, therefore we do not require additional annotations to verify the quality.

\section{Descriptive Behavioral Analysis}

In the descriptive analysis, we employ two distinct prompting strategies. In the first approach, the model is tasked with responding based explicitly on the provided paragraph. In the second approach, no specific instruction is given. Since our models are chat-based, we define and furnish the user message and the system's messages separately. Each model has its own requirements for formatting the two messages, along with an empty initial AI message, in a complete prompt.

\begin{FakepediaExample}[ht]{Descriptive Behavioral Analysis Templates}
\footnotesize
\textbf{User message}
\begin{verbatim}
Question: <query>

Context: <context>

Options:
A) <option_a>
B) <option_b>
\end{verbatim}

\textbf{System message, with explicit instruction}
\begin{verbatim}
Base your response solely on the provided 
context. Respond with 'The correct answer is A'
or 'The correct answer is B', depending on your 
choice. Responses must strictly adhere to this 
format.
\end{verbatim}

\textbf{System message, without explicit instruction}
\begin{verbatim}
Respond with 'The correct answer is A' or 'The 
correct answer is B', depending on your 
choice. Responses must strictly adhere to this 
format.
\end{verbatim}

\end{FakepediaExample}

For the sake of completeness, we also report the same ratios from the models used in the causal tracing experiments (detailed in Section \ref{sec:causal_tracing}) in \Tabref{table:grounded_percentage_CT}. It is important to note that the prompting strategy differs in this case: rather than presenting the two options, we let the model generate the object token. Importantly, the model is not instructed to answer with a grounded answer. When no separate system message is expected or specifiable, we simply concatenate them in a single user message with two new lines in the middle. 

\begin{table}[H]
\centering
\caption{\textbf{Grounding accuracy on Fakepedia from MGCT experiments}}
\begin{tabular}{@{}lcc@{}}
\toprule
& GPT-2 & Llama2 \\
\cmidrule(lr){2-2} \cmidrule(lr){3-3}
Dataset & 1.5B & 7B \\
\midrule
FP & 36.8\% & 58.9\% \\
\midrule
FP-MH & 4.9\% & 5.5\% \\
\bottomrule
\end{tabular}
\label{table:grounded_percentage_CT}
\end{table}

\section{Causal Analysis of the Computation Graph}

\subsection{MGCT Experiments} \label{sec:appendix_causal_tracing}

\xhdr{Experimental setup}
We run MGCT analysis with GPT2-XL and LLaMA-7B on Fakepedia. 
For each model, we partitioned data points into two groups: the instances for which the model's answer is grounded and the ones for which the model's answer is ungrounded. 
When the answer is made of multiple tokens, we consider only the first token, as done previously by \citet{meng2023locating}.

We also share the prompt template for our MGCT experiments.

\begin{FakepediaExample}[ht]{Causal Analysis Template}
\footnotesize
\begin{verbatim}
Context: <context>
Answer: <query>
\end{verbatim}
\end{FakepediaExample}

To execute MGCT, we applied the EOS token as the corruption.

Then, we selected full columns as groups of states to repair simultaneously. This means repairing all states across all layers for a specific token in the Transformer architecture. This grouping requires $K$ mediation analyses to cover all states of the model, where $K$ is the number of tokens in the text (after the first corrupted token). This requires $L$ times less mediation analyses than the previous causal tracing method, where $L$ is the number of layers, but does not give us a measure of effect per layer.

As MGCT effect, we report the percentage of explained in probability between the clean and the corrupted run due to the restoration: 
\begin{equation}
    \frac{P_{*}^{clean(h_k^{(l)})}(o) - P_{*}(o)}{P(o) - P_{*}(o)},
\end{equation}
which is the average indirect effect normalized by $P(o) - P_{*}(o)$, the size of the change in probability to be explained. This makes the MGCT effect comparable across instances with different clean prediction probabilities $P(o)$.

To aggregate different sentences of different lengths, we bucketed tokens into specific categories that we found to be insightful: the first subject token (subj-first), middle subject tokens (subj-middle), the last subject token (subj-last), the first subsequent token (cont-first), middle continuation tokens (cont-middle), and the last token (cont-last). 

\begin{figure*}[htbp]
    \centering
    \begin{subfigure}{0.8\textwidth}
        \centering
        \includegraphics[width=\textwidth]{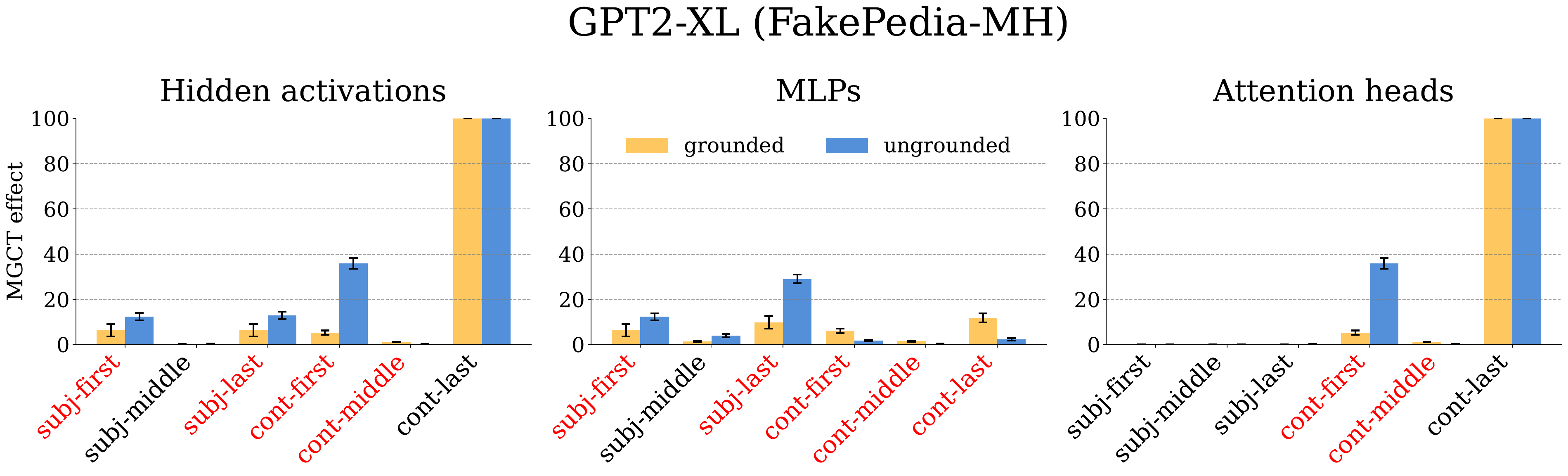}
    \end{subfigure}
    \hfill
    \begin{subfigure}{0.8\textwidth}
        \centering
        \includegraphics[width=\textwidth]{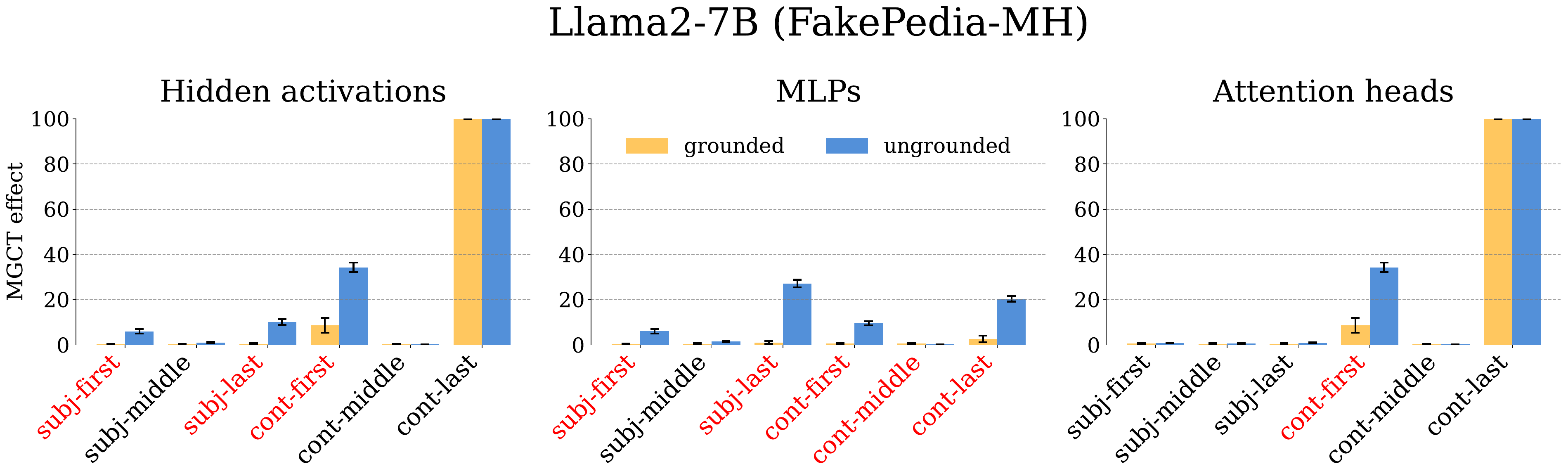}
    \end{subfigure}
    \caption{\textbf{Masked Grouped Causal Tracing analysis on Fakepedia-MH}.This figure illustrates the application of MGCT analysis to GPT2-XL and Llama2-7B on the Fakepedia-MH dataset. We distinguish between instances where the models generated grounded answers and those where they generated ungrounded answers. In the MGCT analysis, we restore full columns together, all states across all layers for a given column at a time, resulting in one effect per token. On the y-axis, we report the percentage of explained change in probability between the clean and corrupted runs due to the restoration of the column. To average across different sequences, we bucketed tokens into subject (subj-) categories and following tokens in the prompt (cont-). Red labels on the x-axis indicate that the difference in MGCT effect between grounded and ungrounded responses is statistically significant based on a t-test with a p-value threshold of 0.01.}
    \label{fig:causal_tracing_Fakepedia_MH}
\end{figure*}

\xhdr{Fakepedia-MH results} We report in \Figref{fig:causal_tracing_Fakepedia_MH} the results from Fakepedia-MH. The observed patterns are similar to those registered with Fakepedia-base. 

\begin{figure*}[htbp]
    \centering
    \begin{subfigure}{0.8\textwidth}
        \centering
        \includegraphics[width=\textwidth]{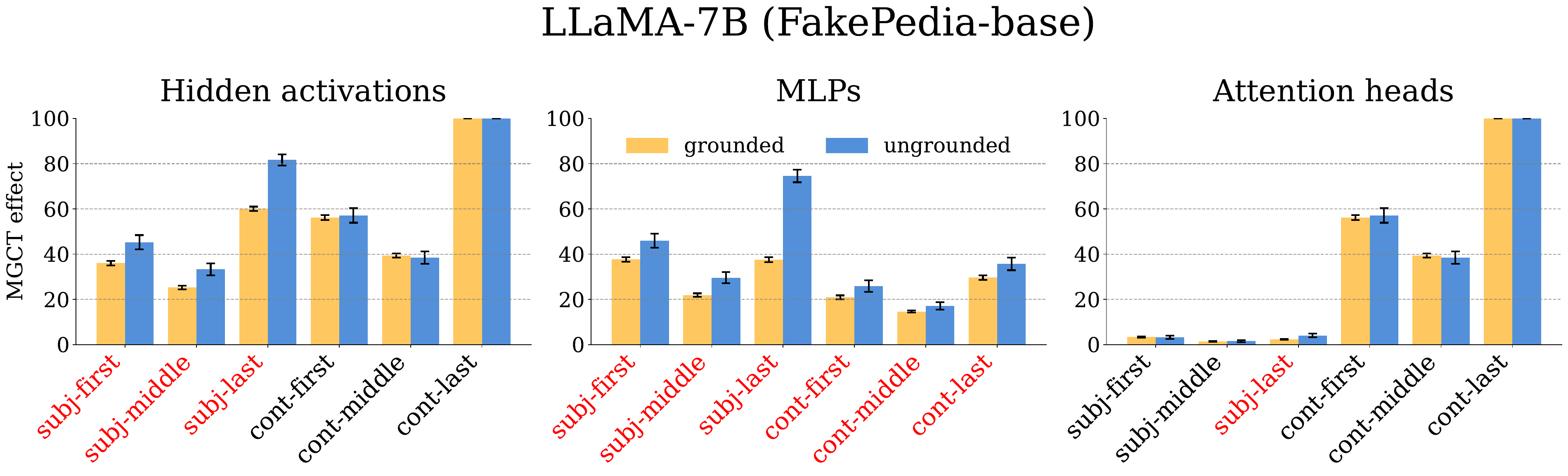}
    \end{subfigure}
    \hfill
    \begin{subfigure}{0.8\textwidth}
        \centering
        \includegraphics[width=\textwidth]{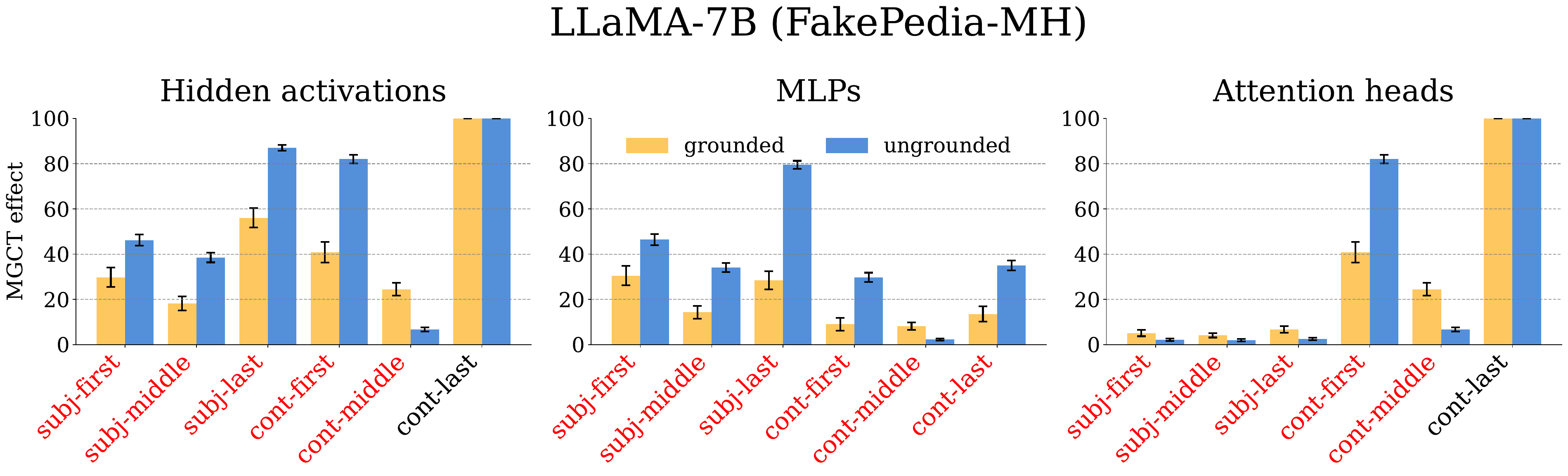}
    \end{subfigure}
    \caption{\textbf{Masked Grouped Causal Tracing analysis on LLaMA}.This figure illustrates the application of MGCT analysis to LLaMA-7B on Fakepedia dataset. We distinguish between instances where the models generated grounded answers and those where they generated ungrounded answers. In the MGCT analysis, we restore full columns together, all states across all layers for a given column at a time, resulting in one effect per token. On the y-axis, we report the percentage of explained change in probability between the clean and corrupted runs due to the restoration of the column. To average across different sequences, we bucketed tokens into subject (subj-) categories and following tokens in the prompt (cont-). Red labels on the x-axis indicate that the difference in MGCT effect between grounded and ungrounded responses is statistically significant based on a t-test with a p-value threshold of 0.01.}
    \label{fig:causal_tracing_llama}
\end{figure*}

\xhdr{LLaMA results} We also share the results from our early experiments with LLaMA in \Figref{fig:causal_tracing_llama}, from both Fakepedia-base and Fakepedia-MH. We find that LLaMA has an overall more distributed structure whether grounded or ungrounded, where more tokens have a high MGCT effect. The prediction is influenced by computation that happens in every part of the Transformer. In comparison, the most important token position for both GPT2 and Llama2 is often the last token.

\section{Automatic Detection of Ungrounded Responses}
\label{sec:appendix_automatic_detector}
The MGCT analysis reveals clear differences in computation patterns between grounded and ungrounded scenarios. We now explore the possibility of using computation patterns to automatically detect whether the model is producing a grounded response or not.

We curate a balanced dataset comprising 4,000 data points of both grounded and ungrounded responses of GPT2-XL on the Fakepedia dataset. This model on this dataset corresponds to the scenario where MGCT plots show minimal differences between grounded and ungrounded instances, i.e., the most challenging scenario for automatic prediction. We partition the dataset into training and test sets in an 80\%-20\% ratio.

The MGCT outputs are transformed into features for a binary classifier, incorporating attention, MLP, and hidden activation effects for each bucket (e.g., subj-first, subj-middle, \etc). This results in a set of 18 features. %

The next step involves training an XGBoost classifier with hyperparameter optimization through cross-validation on the training set. The final classifier is evaluated on the test set, achieving an accuracy of 92.8\%. In an ablation analysis, we remove all features from the MGCT and use only the probabilities derived from the clean and corrupted runs, finding that the model's performance drops significantly to an accuracy of 76.3\%. 

XGBoost allows for easy inspection of feature importance, the total gain a feature contributes across all splits in which it is used. The feature importance analysis of our classier ranks the MLPs on the last subject tokens on top, with a relative importance of $23.4\%$ almost the double of the second best feature. It aligns well with our visual findings with the MGCT analysis in \Figref{fig:causal_tracing_Fakepedia}, hinting that strong MLP effects are predictive of ungrounded answers. 
This is further strong evidence that distinguishing between grounded and ungrounded answers is feasible through an analysis of the computational process alone. Specifically, our efficient MGCT with column group restoration, proves sufficient to robustly predict whether the model is engaged in grounding or not.

\end{document}